# Efficient surrogate modeling methods for large-scale Earth system models based on machine learning techniques

**Dan Lu[1,*], Daniel Ricciuto[2]**

[1]Computational Sciences and Engineering Division, Climate Change Science Institute, Oak Ridge National Laboratory, Oak Ridge, TN, USA;

[2]Environmental Sciences Division, Climate Change Science Institute, Oak Ridge National Laboratory, Oak Ridge, TN, USA;

* Corresponding Author: Dan Lu, lud1@ornl.gov

**Abstract**

Improving predictive understanding of Earth system variability and change requires data-model integration. Efficient data-model integration for complex models requires surrogate modeling to reduce model evaluation time. However, building a surrogate of a large-scale Earth system model (ESM) with many output variables is computationally intensive because it involves a large number of expensive ESM simulations. In this effort, we propose an efficient surrogate method capable of using a few ESM runs to build an accurate and fast-to-evaluate surrogate system of model outputs over large spatial and temporal domains. We first use singular value decomposition to reduce the output dimensions, and then use Bayesian optimization techniques to generate an accurate neural network surrogate model based on limited ESM simulation samples. Our machine learning based surrogate methods can build and evaluate a large surrogate system of many variables quickly. Thus, whenever the quantities of interest change such as a different objective function, a new site, and a longer simulation time, we can simply extract the information of interest from the surrogate system without rebuilding new surrogates, which significantly saves computational efforts. We apply the proposed method to a regional ecosystem model to approximate the relationship between 8 model parameters and 42660 carbon flux outputs. Results indicate that using only 20 model simulations, we can build an accurate surrogate system of the 42660 variables, where the consistency between the surrogate prediction and actual model simulation is 0.93 and the mean squared error is 0.02. This highly-accurate and fast-to-evaluate surrogate system will greatly enhance the computational efficiency in data-model integration to improve predictions and advance our understanding of the Earth system.

# 1   Introduction

Improving predictive understanding of Earth system variability and change requires data-model integration. For example, Bilionis et al. (2015) improved Community Land Model (CLM) prediction of crop productivity after model calibration; Müller et al. (2015) improved the CLM prediction of methane emission after parameter optimization; and Fox et al. (2009) and Lu et al. (2017) improved the terrestrial ecosystem model predictive credibility of carbon fluxes after uncertainty quantification. However, data-model integration methods are usually computationally expensive involving a large ensemble of model simulations, which prohibits their application to complex Earth system models (ESMs) with lengthy simulation time. To reduce computational costs, surrogate modeling is widely used (Razavi et al., 2012; Gong et al, 2015; Ray et al., 2015; Huang et al., 2016, Lu et al., 2018; Ricciuto et al., 2018). The surrogate model, which is a set of mathematical functions, approximates the actual simulation model based on pairs of simulation model input-output samples, and then replaces the simulation model in the data-model integration. As the ESMs evaluation is expensive, it is desired to use a limited number of ESM simulation samples to build an accurate surrogate. As the surrogate model needs to be calculated many times in data-model integration, it is required to build a fast-to-evaluate surrogate. In this study, we use a very few simulation model runs to build an accurate and fast evaluated surrogate system of a large scale problem based on advanced machine learning methods.

In Earth system modeling, we usually need to build a surrogate system of many output variables over large spatial and temporal domains. ESMs tend to be simulated in a regional or global scale with many grid cells for several years, producing a large number of output variables. In addition, ESMs are used to solve versatile scientific problems, so the quantities of interest (QoIs) often change. Moreover, the development of a surrogate requires expensive ESM runs, and a large number of runs are often needed to capture the complex model input-output relationship. Therefore, it is reasonable to build a surrogate system for all possible model outputs to reduce the efforts of rerunning ESMs for a new surrogate development when the QoIs change. In this way, whenever we simulate the outputs in a new site or for additional sites, at a different time or for a longer period, we can simply extract the information of interest from the large surrogate system without spending extra efforts in building new surrogates, which significantly saves the computational costs.

Building and evaluating a surrogate system of a large number of model outputs can be very computationally intensive for almost all the surrogate methods. Polynomials and artificial neural networks are widely used for surrogate modeling (Razavi et al., 2012; Viana et al., 2014). Polynomial methods, such as polynomial regression and radial basis functions, need to solve polynomial coefficients in the surrogate construction and to calculate matrix multiplications in the surrogate evaluation. Using a $p$th-order polynomial to approximate a model with $d$ parameters, $M = (p+d)!/(p!d!)$ coefficients need to be solved, i.e., the number of coefficients increases factorially fast with the parameter size and polynomial order. When $d=40$, a second-order polynomial involves 861 coefficients and a third-order polynomial involves 12341 coefficients. ESMs have many uncertain parameters and a high-order polynomial is usually needed to approximate complex ESMs, which can easily lead to a prohibitive number of model evaluations, up to $\sim 10^5$, necessary to compute the polynomial coefficients. To reduce the computational costs, some regularization techniques such as Bayesian compressive sensing have been used (Sargsyan et al., 2014; Ricciuto et al., 2018). These regularization techniques can use a few samples to solve a large number of coefficients (i.e., an underdetermined system) by iteratively minimizing the L1 norm of the coefficient vector. But they usually perform minimization once for one model output, so for a large model outputs problem, significant computing effort is required. To reduce the computing burden in building polynomial-based surrogates, we need to reduce the output dimensions.

Reducing the model output dimensions also improves computational efficiency in the evaluation of the polynomial-based surrogates. For example, evaluating the third-order polynomial-based surrogate of the model with 40 parameters and 300,000 outputs at 1 parameter sample, we need to calculate two matrix multiplications where matrix A has the size [1, $M$] and B has the size [$M$, Nout] and $M$ =12341 and Nout=300,000. The surrogate evaluation takes about 90 seconds and most time is spent on loading the huge matrix. When Nout reduces to 20, the surrogate evaluation quickly reduces to less than a second. Note that an ESM can easily have more than 40 parameters and more than 300,000 model outputs. Even using the most advanced supercomputers with GPUs, the data storage and loading are still a bottleneck. Thus, reducing model output dimensions is necessary for both fast building and evaluating polynomial-based surrogates.

Neural network (NN) assisted surrogate modeling also suffers from high computational costs when applied to a large-scale problem with many QoIs. To approximate a complex ESM with many outputs, a complicated NN with many wide hidden layers is usually needed to capture the complex relationship between the model inputs and outputs, because each spatial and temporal output variable is driven by different meteorological forcing such as air temperature, humidity, wind speed, precipitation, and radiation. The full connections between nodes in the input layer and the first hidden layer, between nodes of the hidden layers, and between nodes in the last hidden layer and a large number of nodes on the output layer, involve a great amount of NN weights and biases that need to be solved. For the same example discussed above, to approximate the model with 40 parameters and 300,000 model outputs, an NN with two hidden layers and each layer having 100 nodes has over 30 million weights and biases. Calculation of these weights and biases requires many samples to train the NN for a good fit. Each training sample involves one model evaluation. However, ESM simulation is time consuming, which usually takes several hours or days and can be up to months or even years. A limited sample size is not enough to train a deep and wide NN for convergence and a simple NN trained by a small sample size may not capture underlying Earth systems accurately. Thus, reducing model output dimensions is needed to advance the NN-based surrogate modeling. A small output size reduces the width of the output layer and also simplifies the relationship between the model inputs and outputs, so that a simple NN architecture can be appropriate and a small sample size can be sufficient to accurately train the simple NN. In addition, a simple NN can also be fast evaluated with small weight matrix multiplications.

In this work, we propose to use singular value decomposition (SVD) to reduce model output dimensions, so as to improve the computational efficiency in both building and evaluating the surrogates. ESM outputs usually show periodic changes along time and strong correlations between locations, which promises a fast decay of singular values. So, we can use a small number of singular value coefficients to capture a great amount of output information, enabling a significant output dimension reduction. We use the NN for surrogate modeling, because compared to polynomial methods, NNs have shown less difficulty in fitting highly nonlinear and discontinuous functions which are usually observed in ESMs response surfaces. For example, carbon flux state variables, such as gross primary productivity (GPP), are strongly affected by vegetation related parameters. When the parameter samples cause zero vegetation growth, GPP has zero values. Whereas when the parameter samples cause high vegetation growth, GPP has large positive values. This leads to a discontinuous GPP response surface jumping from zeros to nonzeros.

NNs theoretically can fit any functions, but their practical performance strongly depends on the NN's architectures and hyperparameters. NN has many hyperparameters such as the number of layers, number of nodes in each layer, type of activation functions, and learning rate of the stochastic gradient descent optimization. A slight change in the hyperparameter value can result in dramatically different NN performance. Development of a high-performing NN is time-intensive and usually requires trial-and-error tuning by machine learning experts. In this work, we use Bayesian optimization techniques to optimize the NN architecture and hyperparameters so as to produce an accurate NN model for the training data. Bayesian optimization searches the hyperparameter space to iteratively minimize the validation errors of the NN by balancing exploration and exploitation (Shahriari et al., 2016). Researches suggested that Bayesian hyperparameter optimization of NNs is more efficient than manual, random, or grid search with better overall performance on test data and less time required for optimization (Bergstra et al., 2011; Snoek et al., 2012). Bayesian optimization involves a large ensemble of NN fittings and it is a sequential model-based optimization, thus, fast training of the NN models is important. Our proposed SVD method can simplify the NN architecture so as to advance the NN training and improve the Bayesian optimization performance.

In this effort, we propose an SVD-enhanced, Bayesian-optimized, and NN-based surrogate method and aim to build an accurate and fast-to-evaluate surrogate system of a large-scale model using a few model runs, so as to improve computational efficiency in surrogate modeling and thus advance the data-model integration. We apply the method to a simplified land model in the Energy Exascale Earth System Model (sELM) to improve the model predictive capability of carbon fluxes. We build a surrogate system of 42660 model output variables which are annual GPPs at 1422 locations simulated for 30 years. The sELM is a regional-scale terrestrial ecosystem model that simulates terrestrial water, energy, and biogeochemical processes in terrestrial surfaces. Simulation of sELM is important for improving our understanding of ecosystem responses to climate change. However, sELM requires lengthy times for hydrologic and carbon cycle equilibration, and these high computational costs limit the affordable number of simulations in data-model integration thus resulting in poor model performance. The proposed machine learning assisted surrogate method makes the sophisticated data-model integration computationally feasible and promises an improvement of the sELM predictions.

The major contributions of this work are (1) using SVD to reduce model output dimensions so as to improve computational efficiency in both building and evaluating an accurate surrogate of a large-scale ESM; (2) using Bayesian optimization techniques to fast generate an accurate NN-based surrogate; and (3) applying the proposed method to build a large surrogate system of a regional-scale ESM to advance data-model integration. To our knowledge, the method of using SVD to enhance surrogate modeling is novel and we have not seen the application of Bayesian optimization to improve NN-based surrogates in Earth system modeling.

The paper is organized as follows. In section 2, we first describe the sELM, the model parameters and the QoIs we build surrogates for; following that, we introduce the SVD, NNs, and Bayesian optimization methods. In section 3, we apply the methods to the sELM and analyze the surrogate accuracy. In section 4, we discuss strategies to improve surrogate accuracy and investigate our method's performance in the application of these strategies. In section 5, we end this paper by drawing our conclusions.

## 2 Materials and Methods

### 2.1 Description of sELM and related parameters

We developed a simplified version of Energy Exascale Earth System (E3SM) land model (ELM), or sELM, to simulate carbon cycle processes relevant for Earth system models in a computationally efficient framework. This framework allows us to perform large regional ensembles that are computationally infeasible using offline land surface models such as ELM. sELM is a combination of model elements from the Data Assimilation Linked Ecosystem Carbon model (DALEC; Williams et al., 2005) and the Community Land Model version 4.5 (CLM4.5; Oleson et al., 2013). sELM consists of five process-based submodels that simulate carbon fluxes between five major carbon pools using 49 overall parameters. Based on previous sensitivity analysis using ELM (Ricciuto et al., 2018), this study considers the most sensitive eight parameters associated with four out of the five submodels. We summarize all five process-based submodels and their interactions below and in Figure 1.

sELM consists of five major submodels: photosynthesis, autotrophic respiration, allocation, deciduous phenology, and decomposition. Photosynthesis is driven by the aggregate canopy model (ACM) from the DALEC, which itself is calibrated against the soil-plant-atmosphere model (Williams et al., 2005). ACM predicts GPP as a function of carbon dioxide concentration, leaf area index, maximum and minimum daily temperature, and photosynthetically active radiation. Here the GPP predicted by ACM is modified by BTRAN, which reduces GPP when soil water is insufficient to support transpiration. Because sELM does not predict soil moisture, BTRAN is calculated in a full ELM simulation and is fed into sELM as an input. ACM shares one parameter, the leaf carbon to nitrogen ratio (*leaf C:N*), with the autotrophic respiration model and employs an additional parameter, the specific leaf area at the top of the canopy (*slatop*).

The remaining four submodules are based on ELM. The autotrophic respiration model computes the growth and maintenance respiration components and is controlled by four parameters, the *leaf C:N*, the fine root carbon to nitrogen ratio (*froot C:N*), the base rate of maintenance respiration (*br_mr*), and temperature sensitivity for maintenance respiration (*q10_mr*). The allocation model partitions carbon to several vegetation carbon pools following those in ELM: leaves, fine roots, live stem, dead stem, live coarse roots and dead coarse roots. In the allocation model, we only consider one parameter, the fine root to leaf allocation ratio (*froot_leaf*). The deciduous phenology model is used to predict the timing of budbreak and senescence. It considers two parameters, the critical day length to initiate autumn senescence (*crit_dayl*) and the number of accumulated growing degree days needed to initiate spring leaf-out (*crit_onset_gdd*). The last submodel is a decomposition model that simulates heterotrophic respiration and the decomposition of litter into soil organic matter using the converging trophic cascade framework as in the CLM4.5 (Oleson et al., 2013). Because this study focuses on plant carbon uptake, no uncertain parameters are considered in the decomposition model. In sELM, nutrient feedbacks are not represented explicitly, however a constant nitrogen limitation factor is included to downregulate photosynthetic uptake.

The sELM can simulate several carbon state and flux variables as shown in Figure 1 with green shapes. GPP, which represents the total plant carbon uptake, is considered in this study. Here we use sELM to predict annual GPP in deciduous forest systems in the eastern region of the United States for 30 years between 1981-2010. The carbon state variables are spun up to steady state by cycling the GSWP3 input meteorology (Kim et al., 2017) from 1981-2010 for 5 cycles, and the 6<sup>th</sup> cycle is used as the output for our surrogate modeling study. The region of interest covers 1422 land grid cells (locations) as shown in Figure 2. Given 30 outputs at each location (annual values over 30 years), a total of 42660 GPP variables are simulated. The model uses one plant functional type and the phenological drivers such as air temperature, solar radiation, vapor pressure deficit, and $CO_2$ concentration are used as boundary conditions. One regional sELM run takes about 24 hours on a single processor, which although much faster than ELM is still computationally too expensive to be directly used in model-data integration studies. To improve the computational efficiency in generating the sELM simulation samples to develop the surrogate model, we use high performance computing to perform an ensemble of 2000 sELM model simulations in parallel. The 2000 parameter input samples are randomly drawn from the parameter space defined in Figure 3. The numerical ranges of these parameters are designed to reflect their average values and broad uncertainties associated with the temperate deciduous forest plant functional type. The output samples are sELM simulated GPPs at the 1422 locations for 30 years. In the surrogate modeling, part of the 2000 input-output samples are used for developing the surrogate and part of them are used to evaluate the surrogate accuracy, as discussed in section 3.

## 2.2 Efficient surrogate modeling methods

In this section, we introduce our SVD-enhanced, Bayesian-optimized, and NN-based surrogate methods. We first describe the SVD for reducing data dimensionality, then introduce the NN techniques for building a surrogate model, and last depict the Bayesian optimization algorithm for producing a high-performing NN-based surrogate.

### 2.2.1 Singular value decomposition for data compression

We build a surrogate system of model outputs by fitting a data matrix whose columns are output variables and rows are output samples. For a model with 100000 output variables, the columns of this matrix span a 100000-dimensional space. Encoding this matrix on a computer takes quite a lot of memory and evaluating this matrix takes a large number of calculations. We are interested in approximating this matrix with some low-rank matrix but remaining its most information, so as to reduce data transfer and accelerate matrix calculation.

Singular value decomposition (SVD) decomposes a matrix $\mathbf{A}$ with size $m \times n$ into three other matrices, $\mathbf{A} = \mathbf{U}\mathbf{S}\mathbf{V}^T$, where $\mathbf{U}$ is an $m \times m$ orthogonal matrix, $\mathbf{V}$ is an $n \times n$ orthogonal matrix, and $\mathbf{S}$ is an $m \times n$ diagonal matrix saving singular values in descending order on the diagonal. Truncated SVD keeps the $K$ largest singular values and corresponding $K$ column vectors of $\mathbf{U}$ and $K$ row vectors of $\mathbf{V}^T$ to form $\widetilde{\mathbf{A}} = \mathbf{U}_K \mathbf{S}_K \mathbf{V}_K^T$. The $K$-rank matrix $\widetilde{\mathbf{A}}$ has proven to be the best approximation of $\mathbf{A}$ in minimizing the Frobenius norm of the difference between $\mathbf{A}$ and $\widetilde{\mathbf{A}}$ under the constraint of $\mathrm{rank}(\widetilde{\mathbf{A}}) = K$. In addition, the total of the first $K$ singular values divided by the sum of all the singular values is the percentage of information that those singular values contain. For example, if we want to keep 90% of the data information, we just need to compute sums of $K$ largest singular values until we reach 90% of the sum and discard the rest. By dropping all but a few singular values and then recomputing the approximated matrix, the SVD technique compresses the data information and reduces data dimensions. When the matrix $\mathbf{A}$ shows strong correlations between columns (variables), a low-rank matrix $\widetilde{\mathbf{A}}$ can make a very accurate approximation of $\mathbf{A}$.

In this study, we use SVD to reduce training data dimensions. The training data matrix $\mathbf{A}$ [$m$, $n$] for surrogate construction contains model output samples information. $n$ columns are output variables (e.g., the 42660 temporal and spatial GPPs in this work) and *m* rows are the samples of these variables (e.g., the sELM simulation results of the 42660 GPPs for the *m* parameter samples), and usually *n*≫*m* for expensive ESMs with many outputs. In implementation, we first perform truncated SVD to get low-rank matrices $\mathbf{U}_K[m,K]$, $\mathbf{S}_K[K,K]$, and $\mathbf{V}_K^T[K,n]$ with $K \ll n$, we then use the low-dimensional dataset $(\mathbf{V}_K^T \mathbf{A}^T)^T$ with reduced size *m* × *K* as training data to build the surrogate model of the *K* largest singular value coefficients. Next, we evaluate the surrogate model at *q* new data points to get results $\mathbf{Y}_{\text{new}}$ with size $q \times K$. Lastly, we transform the predicted values back to its original size $q \times n$ through $\mathbf{Y}_{\text{new}} \mathbf{V}_K^T$ to obtain the surrogate approximation of the *n* variables at the *q* new data points.

### 2.2.2 Neural networks for surrogate modeling

Artificial neural networks (NNs) consist of fully connected hierarchical layers with nodes which can be flexibly used for function approximation (Yegnanarayana, 2009). The first layer is the input layer and each node in the input layer represents one model input variable. The last layer is the output layer and each node in the output layer represents one model output variable. The layers between input and output layers are hidden layers which are used to approximate the relationship between model inputs and outputs. When the relationship is complex, a complicated NN with many wide hidden layers is usually needed. The input layer first assigns model parameter values to its nodes. Then each node in the first hidden layer takes multiple weighted inputs, applies the activation function to the summation of these inputs, and calculates the node's value. Next, the second hidden layer takes the values on the first hidden layer nodes as inputs and calculates its nodes' values in the same way. This process moves forward till we get values of all nodes in the output layer, i.e., obtaining NN predictions for the given model parameter input values. The nodes in each layer are fully connected to all the nodes in its previous and subsequent layers. Each of these connections has an associated weight and bias. A complicated NN results in a large number of weights. By tuning these weights and biases based on some training data, we improve the NN approximation of the underlying simulation model.

NN uses stochastic gradient descent (SGD) method to optimize its weights and biases (Bottou, 2012). SGD optimizes variables by minimizing some loss function based on the function's gradients to these variables. The loss function is usually defined as the mean squared error (MSE) between the NN predictions and model simulations for the same set of model parameter samples in the training data. SGD iteratively updates the optimized variables at the end of each training epoch. In the process, the learning rate, which specifies how aggressively the optimization algorithm jumps between iterations, greatly affects the algorithm's performance and has to be tuned. A small learning rate will take a long time to reach the optimum causing a slow convergence, whereas a big learning rate will bounce around the optimum causing unstable results and a difficult convergence. Using SGD to optimize a complex NN with many weights requires a great amount of computational efforts and has difficulty in convergence. First, many training data are required to tune a large number of weights. Small training data can easily cause over-fitting, i.e., the NN "perfectly" fits the training data but performs badly on new data, thus deteriorating the NN prediction accuracy. In addition, a large number of weights involve massive matrix calculations in evaluating the loss function, slowing down the training process. Furthermore, a complicated NN has difficulty in convergence and can easily get stuck in local minima. In this work, we use SVD to reduce the model output dimensions, so as to decrease the number of nodes in the output layer and simplify the NN architecture, thus reducing the size of the weights and enabling a reasonable NN training from small training data, and ultimately improving the computational efficiency.

**2.2.3    Bayesian optimization algorithm for NN hyperparameter optimization**

NN involves a lot of hyperparameters that dramatically affect its performance such as the number of layers, the number of nodes in each layer, and the learning rate of the SGD algorithm. Hyperparameter optimization is needed to produce a high-performing NN. This requires optimizing an objective function $f(x)$ over a tree-structured configuration spaces $x \in X$, where some leaf variables (e.g., the number of nodes in the third hidden layer of an NN) are only well defined when branch variables (e.g., a discrete choice of how many layers to use) take particular values. In addition, the optimization not only optimizes discrete and continuous variables, but also simultaneously choose which variables to optimize. When the NN is used for surrogate modeling, the objective function is the NN accuracy of predicting some validation data. In this case, the $f(x)$ does not have a simple closed form but can be evaluated at any arbitrary query point $x$ in the configuration space. For such optimization problem, a sequential search method is needed, besides some inefficient grid search and random search approaches (Bergstra and Bengio, 2012). The sequential search method starts with some random points in the search space, and then iteratively evaluates new points based on NN predictions on previously evaluated points. After $N$ evaluations, we choose the optimal combination of the hyperparameters resulting in the highest NN prediction accuracy. Among the sequential search algorithms, Bayesian optimization is able to take advantage of full information provided by the history of the optimization to improve the search efficiency.

Tree-structured Parzen estimator (TPE) and Gaussian process are two widely used Bayesian optimization algorithms (Shahriari et al., 2016; Bardenet and Kegl, 2010; Niranjan et al., 2010; Snoek et al., 2012). In comparison to the Gaussian process, TPE works well for all types of NN hyperparameter variables, is robust to NN randomization, has a fast calculation and a straightforward implementation without associated hyperparameters specification (Bergstra et al., 2011). In this work, we use the TPE algorithm for NN hyperparameter optimization.

## 3 Results

In this section, we present the results of building the surrogate system of 42660 GPP variables of sELM. First, we demonstrate that our method using SVD can efficiently build and evaluate a large surrogate system by comparing the results with and without application of SVD. We then investigate the influence of NN's architecture on surrogate performance and show that our method using hyperparameter optimization can fast generate an accurate NN. Last, we evaluate surrogate accuracy on the large-scale spatial and temporal GPPs.

We consider three sets of data, the training data for fitting the NN, the validation data to detect overfitting in the NN training and to select the best-performing NN in the hyperparameter optimization, and the test data to evaluate the NN prediction accuracy. Each data set contains pairs of parameter and GPP samples. The parameter samples are randomly drawn from the parameter space defined in Figure 3. To assess the effectiveness of our proposed surrogate method for a small data set, we consider only 20 training data (Figure 3). The validation data is chosen as 0.3 fractions of the training data. The NN model will not train on the validation data but evaluate the loss function on them at the end of each epoch. In each epoch, the training data is shuffled, and the validation data are always selected from the last 0.3 fraction. Precisely, we only use 14 samples to tune NN weights. Attribute to shuffling, these 14 samples can be a different subset from the 20 training data in each epoch, thus we sufficiently explore the limited 20 data information for building the surrogates. We use 1000 test data (Figure 3) to evaluate the NN prediction accuracy, which makes a reasonable assessment of our proposed method within an affordable computational cost. Note that the 1000 test data are not needed for building the surrogates but used to demonstrate the effectiveness and efficiency of our method. When using our method to build the surrogates of the 42660 GPPs, only 20 sELM model simulations are used.

We define the loss function as the mean squared error (MSE) between the NN predictions and the sELM simulations based on the parameter samples for training. We use Adam algorithm (Kingma and Ba, 2015) for stochastic optimization of NN and run it for 800 epochs to minimize the loss function and update NN weights. Adam has been shown a superior stochastic optimization algorithm in training NN (Basu et al., 2018). There is no right answer for the optimal number of epochs. A small number of epochs could result in underfitting and a large number of epochs may lead to overfitting. Here we consider a large number of epochs and in the meantime use early stopping to avoid overfitting. During the training, when there is no improvement of loss functions for the validation data in 100 epochs, we stop the training and choose the weights at the epoch resulting in the smallest loss function of the validation data as the optimal weights and the associated NN as the best trained NN under the given setting.

We then use the trained NN to predict the 1000 test data and compare the predictions with the corresponding sELM simulation results to evaluate the NN accuracy. We define two metrics for evaluation, the MSE and the coefficient of determination. The MSE computes the expected value of the squared prediction errors; the small the MSE value is, the better the prediction. The coefficient of determination, also called $R^2$ score, measures how well the unobserved data are likely to be predicted by the NN model. Denote $\hat{y}_i$ as the NN prediction of the *ith* sample and $y_i$ as the corresponding sELM simulation, the $R^2$ score estimated over $N_s$ samples is defined as

$R^2 = 1 - \frac{\sum_{i=1}^{N_s}(y_i - \hat{y}_i)^2}{\sum_{i=1}^{N_s}(y_i - \bar{y}_i)^2}$ , where $\bar{y} = \frac{1}{N_s}\sum_{i=1}^{N_s} y_i$. Best possible value of $R^2$ score is 1.0, indicating that the NN can perfectly predict the test data. $R^2$ score can be negative indicating the model is arbitrarily poor. A constant model gets a $R^2$ score of 0.0. Compared to MSE, the $R^2$ score considers the variability of the data which provides a more reasonable measure.

## 3.1 SVD reduces data dimensionality and improves surrogate efficiency

We consider two scenarios when building the surrogate system of the 42660 GPP outputs; Case I: building the surrogates of reduced data after SVD, and Case II: building the surrogates of all GPPs directly. In Case I, we first apply SVD to reduce the training data dimensionality, then build surrogates of the singular value coefficients, and last transfer the surrogate system back to the original QoIs (i.e., the 42660 GPP variables).

The goal of this study is to develop a surrogate method that builds an accurate surrogate system with small training data, so as to reduce the computational costs in simulating the expensive ESMs. To demonstrate the effectiveness and efficiency of our method, we compare the surrogate performance of the two cases in predicting the 1000 test data from two aspects: (1) for the same number of training data, the predictive accuracy of the two surrogates, and (2) the number of training data used to achieve the similar predictive accuracy.

Figure 4 shows the singular value decay of decomposition of the training data matrix having 20 samples and 42660 GPP variables. The figure indicates that the singular values decay very fast. The first 2 singular values drop about 1 magnitude, and the first 5 singular values can capture 97% information of the training data matrix. To choose a suitable number of singular value coefficients (Nsvd) to compress the training data and build a surrogate for, we consider a series of Nsvds, where Nsvd=1, 5, 10, 15, and 20, and investigate their impact on NN performance. To make a fair comparison, the same NN architectures are used for all Nsvd cases. We consider a simple NN with 2 hidden layers and each hidden layer has 10 nodes. Figure 5 shows the prediction performance of the NNs based on the 20 training data. The figure indicates that with considering only 1 singular value coefficient, the averaged MSE of the predictions is about 0.053, and the NN model can fit the sELM simulation data well with the $R^2$ score of 0.83. When 5 singular value coefficients are considered, the NN prediction accuracy improves with the MSE of 0.02 and the $R^2$ score of 0.93. After Nsvd=5, the MSE and $R^2$ score have minor changes, suggesting that for the limited 20 training data, Nsvd=5 is a good choice to compress the GPPs and build a surrogate for. At this time, the surrogate error becomes dominant compared to the SVD approximation error and including more than 5 singular value coefficients would barely improve the NN prediction unless more training data are included to reduce the surrogate error. In the following, we consider Nsvd=5 in Case I and compare its surrogate prediction performance with Case II which builds surrogates for all GPPs directly.

In Case I, our method is able to use 20 training data to build a highly accurate surrogate of 42660 GPP variables with a small MSE of 0.02 and a high $R^2$ score of 0.93. The detailed NN performance is explained in Figure 6(a) where the training and validation loss decays in building the surrogates of the 5 singular value coefficients are plotted. The figure indicates that the loss functions of the two data sets have similar decay, decreasing dramatically at the first 10 epochs and then slowly decreasing to the end of training. The closely overlapped two lines in Figure 6(a) suggest that the trained NN captures the relationship between sELM inputs and outputs pretty well and can give reasonable predictions of GPPs for a given parameter sample.

To make a fair comparison, we use the same NN architecture in Case II as in Case I except that the output layer of NN in Case II has all the 42660 GPPs and the output layer in Case I has only 5 singular value coefficients. Figure 6(b) indicates that the simple NN with 20 hidden nodes is not sophisticated enough to capture the complex relationship between the 8 inputs and 42660 outputs. As we can see in Figure 6(b), both training and validation losses are relatively high suggesting an underfitting. The validation loss is always larger than the training loss suggesting that the fitted NN does not generalize well and may result in poor performance in predicting new data. Figure 7 shows $R^2$ scores of Case II in predicting the 1000 test data. The figure indicates that the simple NN trained by 20 data in Case II has a very poor prediction accuracy with the $R^2$ score of only 0.05, close to a constant model's performance with a zero $R^2$ score. However, with the same NN trained by the same 20 data, our SVD-based surrogate method can achieve a high prediction accuracy with the $R^2$ score of 0.93. This demonstrates our method's capability in using a few training samples to build an accurate surrogate model, greatly reducing the computational costs in generating the expensive model simulation data.

On the other hand, the poor performance in Case II suggests that a wider and deeper NN is needed when we consider the large outputs directly. We thus increase the nodes of each hidden layer to 100 and use this complex NN with total 200 hidden nodes to approximate the relationship of the 8 inputs and 42660 outputs in Case II. This complex NN blows up its parameters (including weights and biases) to 4.3 million from 255 in Case I. To fit this wide NN and calibrate its large parameters, 20 training data are way too small to get a reasonable fit. No matter how we adjust the NN hyperparameters, we cannot get a stable solution in training. We then increase the training data to 50, Figure 6(c) shows that the increased data greatly decrease the training and validation losses and the validation loss is slightly higher than the training loss, implying a good fit. Figure 7 indicates that the complex NN with 200 hidden nodes trained by 50 data in Case II significantly improves the prediction accuracy with the $R^2$ score of 0.73. However, Case II's predictive performance is still worse than Case I which has the $R^2$ score of 0.93. We keep increasing the training data (Ntrain) to 100 and 200 in Case II. Figure 6(d) and (e) indicate that the increase of training data brings the validation loss closer and closer to the training loss making the fitted NN represent the underlying sELM better and better. Figure 7 shows that the nicely fitted NNs trained by large Ntrains lead to a high prediction accuracy. With Ntrain=100, the $R^2$ score is about 0.89, and with Ntrain=200, the $R^2$ score is up to 0.95. However, compared to Case I using 20 training data to get predictive $R^2$ score of 0.93, Case II uses near 200 data to get the similar accuracy, increasing 10-fold computational costs. Note that, each training data involves one sELM simulation and one regional sELM run takes about 24 hours on one processor. Thus, our SVD-based surrogate method greatly improves computational efficiency in the accurate surrogate modeling.

Our method, in the means of simplifying NN architecture through data compression, not only reduces the training data but also decreases the training time. Using 20 data to train a simple NN with 255 parameters, our method takes about 4 seconds. In comparison, the traditional surrogate method without data compression spends a great effort in training the complex NN with 4.3 million parameters. As shown in Figure 7, Case II takes 270 seconds to fit the NN based on 50 training data and 967 seconds for the 200 training data, showing a linear increase in computing time. The long training time leads to high computational costs in NN hyperparameter optimization where massive NN training are involved in searching the wide hyperparameter space for a high-performing NN model, as discussed in the following section 3.2.

### 3.2 NN's hyperparameter optimization improves surrogate accuracy

NN has a large number of hyperparameters. Here we adjust 5 hyperparameters and use Case I to investigate their influence on surrogate prediction accuracy. The 5 hyperparameters are, the number of hidden layers (L) where we consider the most 3 hidden layers, the number of nodes in hidden layer 1 (N1), in hidden layer 2 (N2), and in hidden layer 3 (N3), and the learning rate (lr) of Adam optimization algorithm. We consider the following choices: L={2, 3}, N1={10,

20, 40, 60, 80, 100}, N2={10, 20, 40, 60, 80, 100}, N3={0, 10, 20, 40, 60, 80, 100}, and lr=U[0.001, 0.1]. The first four hyperparameters are discrete variables and the last one, lr, is a continuous variable with uniform distribution. The choice of L determines the selection of N3 showing a tree-like structure. We use tree-structured Parzen estimator (TPE) to search the 5 hyperparameter space and find a set of values that gives the best-performing NN. We fix the activation function as ReLU (Agarap, 2018) which has been widely used and shown to produce good NN predictions.

    We use TPE to evaluate 100 sets of hyperparameters and the one giving the best validation score, i.e., the smallest MSE on validation data, is chosen as the optimal hyperparameters. Results indicate that the combination of N1=10, N2=10, N3=0, and lr=0.08 gives the best validation score. To investigate the impact of hyperparameters on NN prediction accuracy, we show the 100 sets of hyperparameters and their resulting $R^2$ scores in predicting the 1000 test data in Figure 8. The figure indicates that different hyperparameter values result in dramatically different NN performance. The prediction $R^2$ scores range from 0.66 to 0.93 where 32 hyperparameter sets have the $R^2$ scores over 0.90. The selected optimal NN producing the smallest MSE on the validation data also gives the best prediction performance on the test data with the $R^2$ score of 0.93. It is desired that the best NN model chosen by validation data gives the best predictions, however, in practice it is not always the case, especially when the prediction data deviates a lot from the validation data. Extrapolation is always a difficulty in surrogate modeling and several researches are going on to improve the extrapolation accuracy (Gal, 2014).

    Although NNs perform significantly different with different combination of hyperparameters, the TPE algorithm can efficiently find the high-performing NNs based on previous samples information. As shown in Figure 8, good-performing NNs prefer simple

architectures with 2 hidden layers, e.g., most blue lines have N3 of 0. After TPE finds a good architecture of N1=10 and N2=10, it samples around this architecture in the hyperparameter space to fine tune the learning rate till finds the most suitable lr of 0.08. This work considers 5 hyperparameters with limited choices, increasing the dimensions and possible choices of the hyperparameters would make the search more thorough and could produce a better-performing NN. Our surrogate method with SVD can accelerate the optimization process by reducing the NN training time.

**3.3 Evaluation of surrogate accuracy on large-scale spatial and temporal data**

We, using only 20 expensive sELM runs, fast build an accurate surrogate system of 42660 GPPs at 1422 locations for 30 years. Therefore, for a data-model integration problem with the QoIs within the spatial and temporal ranges, we can directly extract the information of interest from the surrogate system to advance the analysis. The best-performing NN generated from our method gives an overall accurate prediction of the 42660 GPPs with averaged MSE of 0.02 and $R^2$ scores of 0.93. When using the subset of the surrogate system for data-model integration studies, it is desired to analyze the surrogate accuracy at individual locations for specific times.

Figure 9 shows averaged $R^2$ scores over 30 years at 1422 locations. The figure indicates that the surrogate accuracy is not uniformly good for all the locations. We observe that most locations have $R^2$ scores above 0.9 with the best $R^2$ score of 0.96, and about 100 locations have $R^2$ scores below 0.90 with the smallest $R^2$ score of 0.79. We highlight the locations having zero GPP simulations in blue circles and find that these locations generally have poor predictions with low $R^2$ scores. Connecting to Figure 2 where we label the locations in column-wise from south to north and from west to east, we identify that those locations with zero GPPs are mostly located in the north where the temperature is relatively low and annual GPPs tend to be zero for parameter samples.

We pick 3 locations to closely evaluate the surrogate accuracy (Figure 9). Location 1046 has the best prediction with the highest $R^2$ score, location 1345 has the worst prediction accuracy, and location 428 performs best among the locations with zero GPP simulations. Figure 10 shows annual GPP simulations based on sELM and NN-based surrogate in evaluating the 1000 test data for 30 years at the 3 locations. It can be seen that NN has difficulty in fitting zero GPP data. At location 1046 where the annual GPPs are relatively high with positive values, NN produces a great fit with a high $R^2$ score of 0.96 and a small MSE of 0.013. Location 1046 (Figure 2) is close to the lake where the variance in atmospheric drivers (e.g., temperature) is moderated. This reduced variance leads to a smooth response surface of GPP for which NN can easily build an accurate surrogate. In contrast, location 1345 has a large number of simulated GPPs less than 1.0 including many zero GPPs. NN shows difficulty in predicting these small GPPs resulting in a relatively poor performance with the $R^2$ score of 0.79. Location 1345 is sitting in the north and has the lowest mean annual temperature, so the most parameter samples cause low vegetation growth and small GPP values. Moreover, location 1345 is far away from the lakes and has a large variation in atmospheric drivers. Since this location has a climate that is at the extreme end of the range for deciduous forests, the model response is expected and reasonable. However, this leads to a strong nonlinear response surface that casts difficulty in surrogate modeling. In comparison, although location 428 is located in the north with some small GPPs including zero values, it is also close to the lake which has a small variance in the atmospheric drivers. Thus, the NN prediction performance in location 428 is not bad with the $R^2$ score of 0.91.

Figure 11 plots the averaged $R^2$ scores over all locations for 30 years. The $R^2$ scores have small fluctuations between 0.93 and 0.94, displaying a uniformly good fit among the simulated years. So, when using the surrogate model at any specific year for a data-model analysis, we should be able to obtain a good approximation. In this study, we are considering annual GPPs. Although the variation of atmospheric drivers between years has an impact on surrogate accuracy, its influence is less strong compared to monthly GPPs, so a uniformly good fit among years is expected.

Building a surrogate of the discontinuous response surface, e.g., vegetation turns from alive to dead representing as the GPP jumps from nonzero to zero, is a difficulty for almost all the state-of-the-art surrogate methods. Nevertheless, NNs, attribute to the layered architecture and the nonlinear activation function, usually show better performance compared to other surrogate approaches. To improve the surrogate accuracy for strong nonlinear and discontinuous problems, one strategy is using physics-informed domain decomposition methods to build surrogate models separately in different response surface regimes. This strategy requires the surrogate methods strongly connecting to the simulation model, and the methods are generally problem-specific requiring experts' interaction. Another strategy is increasing the training data to explore complex problems. This strategy requires an increase in computational costs for extra expensive model simulations. In the following section 4, we investigate these two strategies and discuss their influence on surrogate accuracy.

## 4 Discussion

ESMs are complex whose response surfaces always display strong nonlinearity and discontinuity, casting a challenge to surrogate modeling. In this section, we consider the strategies of physics-informed learning and increase of training data to improve the surrogate accuracy. We conduct two corresponding experiments to investigate our method's performance in application of these two strategies. In experiment I, we divide the parameter space into two parts producing zero GPPs and nonzero GPPs, and we use 20 training data to build surrogates of the 42660 GPPs in the regime generating nonzero GPP samples. In experiment II, we build the surrogates of the 42660 GPPs in the original parameter domain (Figure 3), but with increasing training data of 200 and 1000.

We use the results of Case I as a baseline to investigate our method's performance in the two experiments. Figure 12 shows averaged $R^2$ scores over 30 years at the 1422 locations in experiment I. The figure indicates that without zero GPPs our method can produce a very accurate surrogate at all locations with a uniformly high $R^2$ score of 0.98. Building the surrogates in the subdomain without zero GPPs not only significantly improves the prediction accuracy in locations originally having poor fit in Case I, but also further improves the prediction accuracy in locations which already have a good fit in Case I. For example, the $R^2$ score is dramatically improved from 0.79 to 0.97 at location 1345, from 0.96 to 0.99 at location 1046, and from 0.91 to 0.98 at location 428. As shown in Figure 13, the NN almost perfectly reproduces sELM simulations at these 3 locations. Experiment I indicates that physics-informed domain decomposition can be a good strategy to improve surrogate accuracy. For smooth problems (e.g., no sharp jumps from non-zeros to zeros in response surfaces), our method can build a very accurate surrogate model based on a few training data.

Figure 14 shows averaged $R^2$ scores over 30 years at 1422 locations based on 200 and 1000 training data in experiment II. The figure indicates that an increase of training data greatly enhances NN prediction accuracy. Adding 10 folds additional data from Ntrain=20 to Ntrain=200, the overall $R^2$ score improves from 0.93 to 0.98; further increasing Ntrain to 1000, the averaged $R^2$ score is up to 0.993 with the worst value of 0.96. Although we observe similar nonuniform performance among locations in Figure 14 as in Figure 9, where the locations with zero GPPs have smaller $R^2$ scores than others, increasing Ntrain significantly improves the accuracy at all locations, especially those originally having poor fits in Case I. For example, when Ntrain=200, most blue-circled locations have $R^2$ scores above 0.95 and for Ntrain=1000, the $R^2$ scores at these blue-circled locations are above 0.985 in comparison to the values below 0.9 when Ntrain=20. In the examination of the 3 individual locations by comparing Figure 10 and Figure 15, we see that at the location of 1046, an increase of Ntrain enables the NN to perfectly predict sELM simulations with negligible MSEs. Even for the location 428 with zero GPPs, more training data can capture the discontinuous behavior better with $R^2$ score of 0.99 and MSE of 0.003 when Ntrain=1000. The worst location happens at 1345 for all cases due to its highly changed atmospheric drivers. Even so, the increase of Ntrain can still dramatically enhance the NN's capability in simulating the difficult response surface. Experiment II indicates that increasing training data is able to significantly improve the surrogate accuracy. Our method scales well with the increase of training data and greatly improves prediction accuracy as Ntrain increases.

The analysis of the two experiments suggests that our method is data-efficient for continuous problems. To improve the surrogate accuracy in discontinuous and highly nonlinear problems, we can use the physical-informed domain decomposition to focus on the continuous and smooth regions of the response surface. If the discontinuity is the inherent feature of the underlying function that we need to surrogate, an increase of training data would be a good solution for surrogate accuracy improvement.

Having built a surrogate system of many GPP variables over large spatial and temporal domains provides great flexibility and possibility for subsequent predictive analytics tasks. For example, the surrogate model can be used for analyzing sensitivities of model parameters to any set of spatial and temporal GPP variables, and for parameter optimization and uncertainty quantification based on a single-site or multiple-site, a single-year or multiple-year GPP observations using any defined objective functions. In addition, with the newly collected observations from additional sites or further time periods, we can use the same surrogate system for analysis as long as the QoIs are within the surrogate simulation ranges. In the future study, we will pursue the data-model integration using the constructed surrogate system.

## 5 Conclusions

In this work, we develop an SVD-enhanced, Bayesian-optimized, and NN-based surrogate method to improve the computational efficiency of large-scale surrogate modeling, so as to advance model-data integration studies in Earth system model simulations. Our method is data efficient in the fact that only 20 model simulations are needed to build an accurate surrogate system. This is a promising result because large Earth system model ensembles are always computationally infeasible, and 20 is a reasonable and affordable number of simulations to consider. In addition, our method is general purpose and can be efficiently applied to a wide range of Earth system problems with different spatial scales (local, regional, or global) at different simulation periods. It is super effective for smooth problems and scaled well for highly nonlinear and discontinuous problems.

We apply our surrogate method to a regional ecosystem model. The results indicate that using only 20 model runs, we can build an accurate surrogate system of 42660 spatially- and temporally-varied GPPs with the $R^2$ score of 0.93 and MSE of 0.02. For locations with robust vegetation growth across the ensemble, our method can almost perfectly predict the model simulations with the $R^2$ score of 0.96. For locations with low vegetation growth for some parameter samples and large variation in atmospheric drivers that cause discontinuous response surfaces, using physics-informed domain decomposition or the increase of training samples, our method can produce accurate predictions with the $R^2$ score of 0.97 and 0.96, respectively. This application demonstrates our method's capability in accurately reproducing expensive model simulations based on a few parallel model runs.

**Data availability**

All the data used in this study are model simulation data, which can be generated by running the sELM.

**Code availability**

sELM is presented in its 1.0 version, which is realized in the Python language. It is an open-use computer code which can be accessed freely from https://github.com/dmricciuto/OSCM_SciDAC/tree/master/models/simple_ELM. The source code of surrogate modeling using machine learning techniques can be provided upon request via lud1@ornl.gov.

**Author contribution**

Dan Lu developed the methods and carried them out. Daniel Ricciuto developed the model code and performed the model simulations. Dan Lu prepared the manuscript with contributions from all coauthors.

**Acknowledgments**

Primary support for this work was provided by the Scientific Discovery through Advanced Computing (SciDAC) program, funded by the U.S. Department of Energy (DOE), Office of Advanced Scientific Computing Research (ASCR) and Office of Biological and Environmental Research (BER). Additional support was provided by BER's Terrestrial Ecosystem Science Scientific Focus Area (TES-SFA) project. The authors are supported by Oak Ridge National Laboratory, which is supported by the DOE under Contract DE-AC05-00OR22725.

**References**

Agarap, A. F. M.: Deep learning using Rectified Linear Units (ReLU), https://arxiv.org/pdf/1803.08375, 2018.

Bardenet, R. and Kegl, B.: Surrogating the surrogate: accelerating Gaussian Process optimization with mixtures, In ICML, 2010.

Basu A., De, S., Mukherjee, A., and Ullah, E.: Convergence guarantees for rmsprop and adam in nonconvex optimization and their comparison to nesterov acceleration on autoencoders, arXiv preprint arXiv:1807.06766, 2018.

Bergstra J. S., Bardenet, R., Bengio, Y., and Kegl, B.: Algorithms for hyperparameter optimization, NIPS 24, 2546-2554, 2011.

Bergstra J., and Bengio, Y.: Random search for hyper-parameter optimization, Journal of Machine Learning Research, 13(1): 281-305, 2012.

Bergstra, J., Yamins, D., and Cox, D. D.: Hyperopt: A Python library for optimizing the hyperparameters of machine learning algorithms, In Proceedings of the 12th Python in Science Conference, 13-20, 2013.

Bottou, L.: Stochastic gradient descent tricks, Neural networks: tricks of the trade: 2$^{nd}$ edition, Springer Berlin Heidelberg, 2012.

Bilionis, I., Drewniak, B. A., and Constantinescu, E. M.: Crop physiology calibration in the CLM, Geosci. Model Dev., 8, 1071-1083, 2015.

Fox, A., Williams, M., Richardson, A. D., Cameron, D., Gove, J. H., Quaife, T., Ricciuto, D., Reichstein, M., Tomelleri, E., Trudinger, C. M., and Van Wijk, M. T.: The REFLEX project: Comparing different algorithms and implementations for the inversion of a terrestrial ecosystem model against eddy covariance data, Agric. For. Meteorol., 149, 1597-1615, 2009.

Gong, W., Duan, Q., Li, J., Wang, C., Di, Z., Dai, Y., Ye, A., and Miao, C.: Multiobjective parameter optimization of community land model using adaptive surrogate modeling, Hydrol. Earth Syst. Sci., 19, 2409-2425, 2015.

Huang, M., Ray, J., Hou, Z., Ren, H., Liu, Y., and Swiler, L.: On the applicability of surrogate-based Markov chain Monte Carlo-Bayesian inversion to the Community Land Model: Case studies at flux tower sites, J. Geophys. Res. Atmos., 121, 7548-7563, doi:10.1002/2015JD024339, 2016.

Kim, H.: Global Soil Wetness Project Phase 3 Atmospheric Boundary Conditions (Experiment 1). Data Integration and Analysis System (DIAS), https://doi.org/10.20783/DIAS.501, 2017.

Kingma, D. P., and Ba, J.: Adam: a Method for Stochastic Optimization, International Conference on Learning Representations, 1-13, 2015.

Lu, D., Ricciuto, D., Walker, A., Safta, C., and Munger W.: Bayesian calibration of terrestrial ecosystem models: a study of advanced Markov chain Monte Carlo methods, Biogeosciences, 14, 4295-4314, 2017.

Lu, D., Ricciuto, D., Stoyanov, M., and Gu, L.: Calibration of the E3SM land model using surrogate-based global optimization. Journal of Advances in Modeling Earth Systems, 10. https://doi.org/10.1002/2017MS001134, 2018.

Müller, J., Paudel, R., Shoemaker, C. A., Woodbury, J., Wang, Y., and Mahowald, N.: CH4 parameter estimation in CLM4.5bgc using surrogate global optimization, Geosci. Model Dev., 8, 3285-3310, 2015.

Niranjan, S., Krause, A., Kakade, A., and Seeger, M.: Gaussian process optimization in the bandit setting: No regret and experimental design. In Proceedings of the 27th International Conference on Machine Learning, 2010.

Oleson, K. W., et al.: Technical description of version 4.5 of the Community Land Model (CLM). (NCAR Tech. Note NCAR/TN-5031STR, 420 pp. Boulder, CA: National Center for Atmospheric Research, https://doi.org/10.5065/D6RR1W7M, 2013.

Ray, J., Hou, Z., Huang, M., Sargsyan, K., and Swiler, L.: Bayesian calibration of the Community Land Model using surrogates, SIAM/ASA J. Uncertain. Quantif., 199-233, doi:10.1137/140957998, 2015.

Razavi, S., Tolson, B. A., and Burn, D. H.: Review of surrogate modeling in water resources, Water Resour. Res., 48, W07401, doi:10.1029/2011WR011527, 2012.

Ricciuto, D., Sargsyan, K., and Thornton, P.: The impact of parametric uncertainties on biogeochemistry in the E3SM land model. Journal of Advances in Modeling Earth Systems, 10, 297-319, 2018.

Sargsyan, K., Safta, C., Najm, H. N., Debusschere, B., Ricciuto, D. M., and Thornton, P.E.: Dimensionality reduction for complex models via Bayesian compressive sensing, Int. J. Uncert. Quant., 4, 63-93, 2014.

Shahriari, B., Swersky, K., Wang, Z., Adams, R. P., and de Freitas, N.: Taking the Human Out of the Loop: A Review of Bayesian Optimization, Proc. IEEE 104 (1): 148-175, doi:10.1109/jproc.2015.2494218, 2016.

Snoek, J., Larochelle, H., and Adams, R. P.: Practical Bayesian optimization of machine learning algorithms, in 26th Annual Conference on Neural Information Processing Systems, 2960-2968, 2012.

Williams, M., Schwarz, P. A., Law, B. E., Irvine, J., and Kurpius, M.: An improved analysis of forest carbon dynamics using data assimilation, Global Change Biol., 11, 89-105, 2005.

Viana, F. A., Simpson, T.W., Balabanov, V., and Toropov, V.: Metamodeling in multidisciplinary design optimization: How far have we really come?, AIAA J., 52(4), 670-690, 2014.

Yegnanarayana B.: Artificial neural networks, PHI Learning Pvt. Ltd, 2009.

**List of Figures**

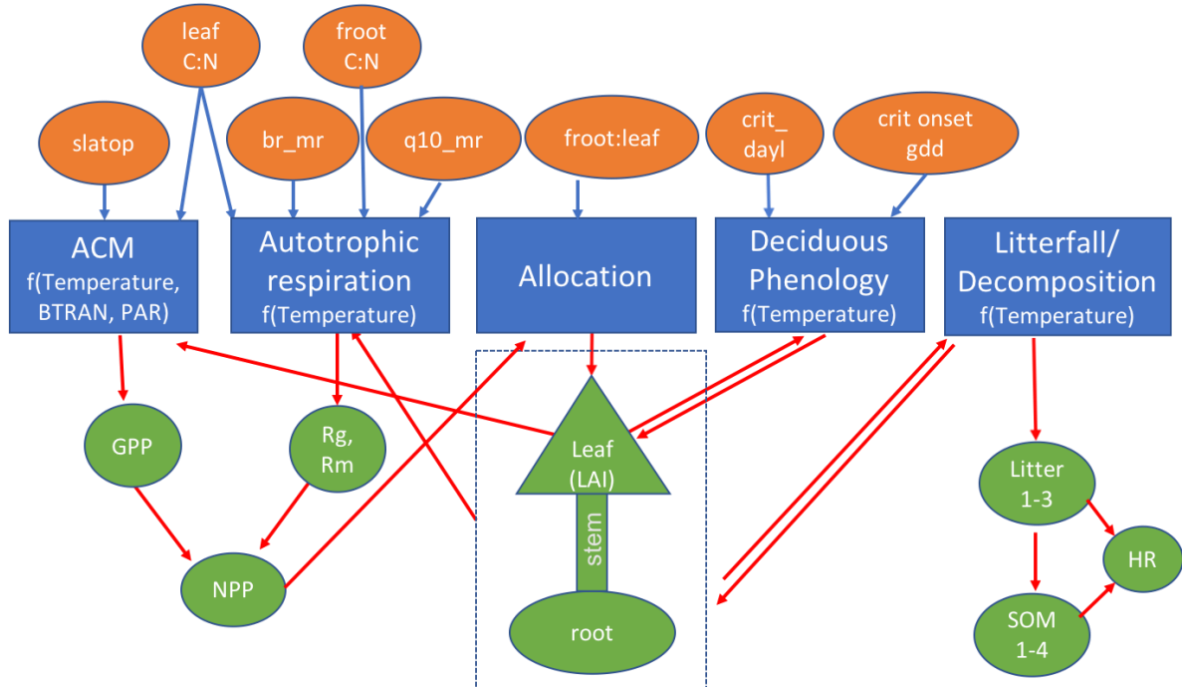

Figure 1. Schematic of sELM, where processes are shown using blue boxes with dependencies on environmental data, 8 uncertain parameter inputs are listed in orange ovals, and model state variables are indicated by green shapes. Parameters are input to one or more processes as indicated by blue arrows. Model state variables may be outputs for some processes and input for other processes as indicated by red arrows.

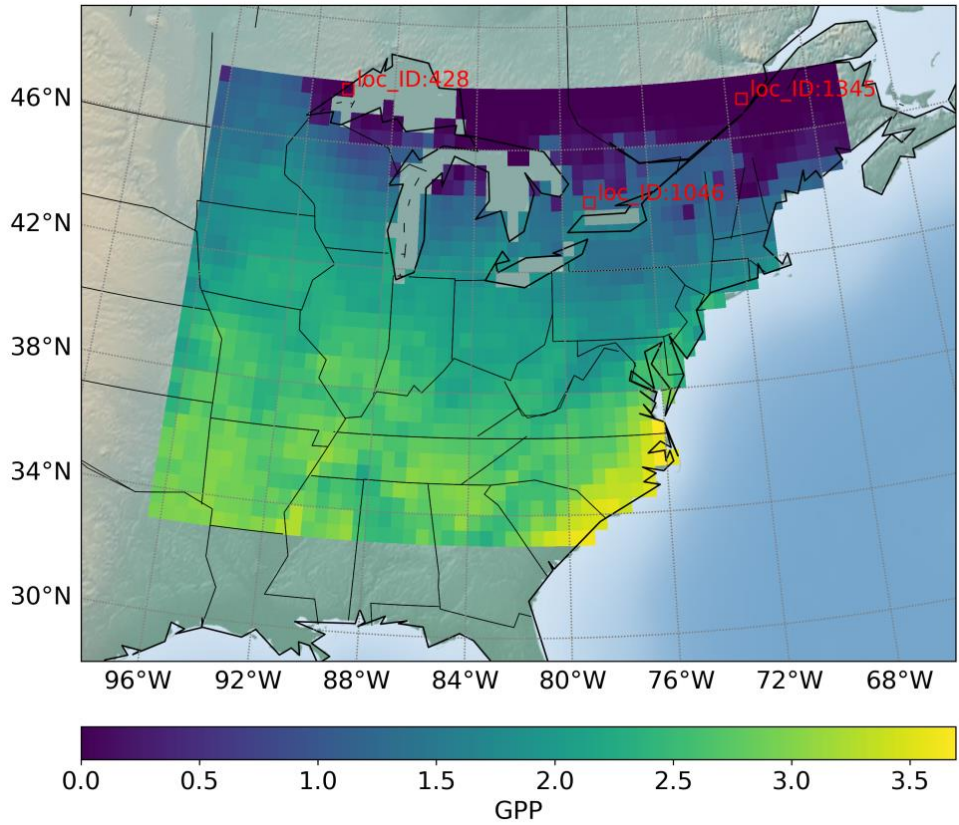

Figure 2. Locations of interest for which we build surrogates of GPP (gC/m^2/day) variables; total 1422 locations are considered. The figure shows the sELM simulated annual GPP based on one parameter sample.

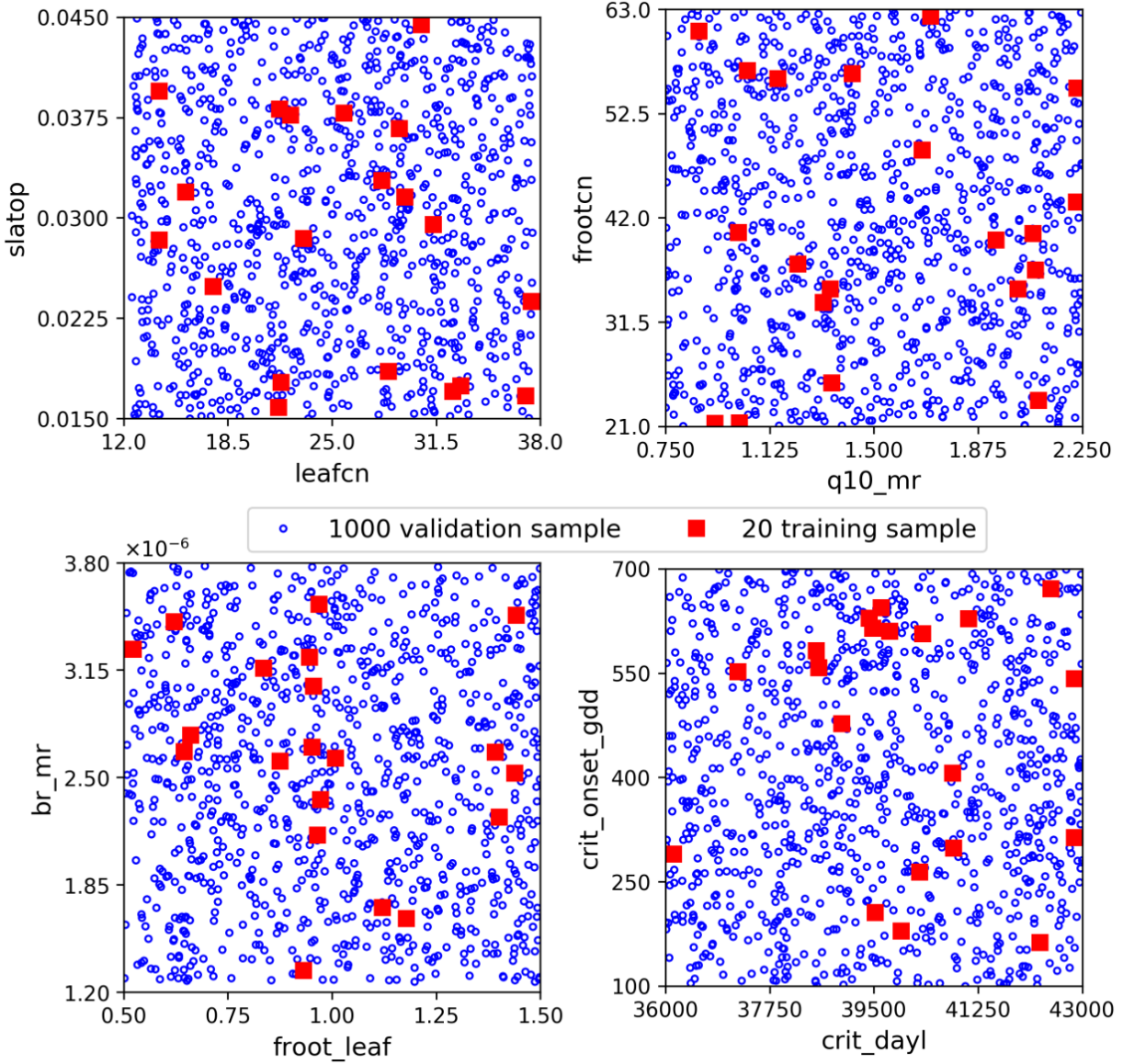

Figure 3. We consider 8 uncertain parameter inputs whose ranges are shown as axis limits. The 20 training and 1000 test data are randomly drawn from the parameter space.

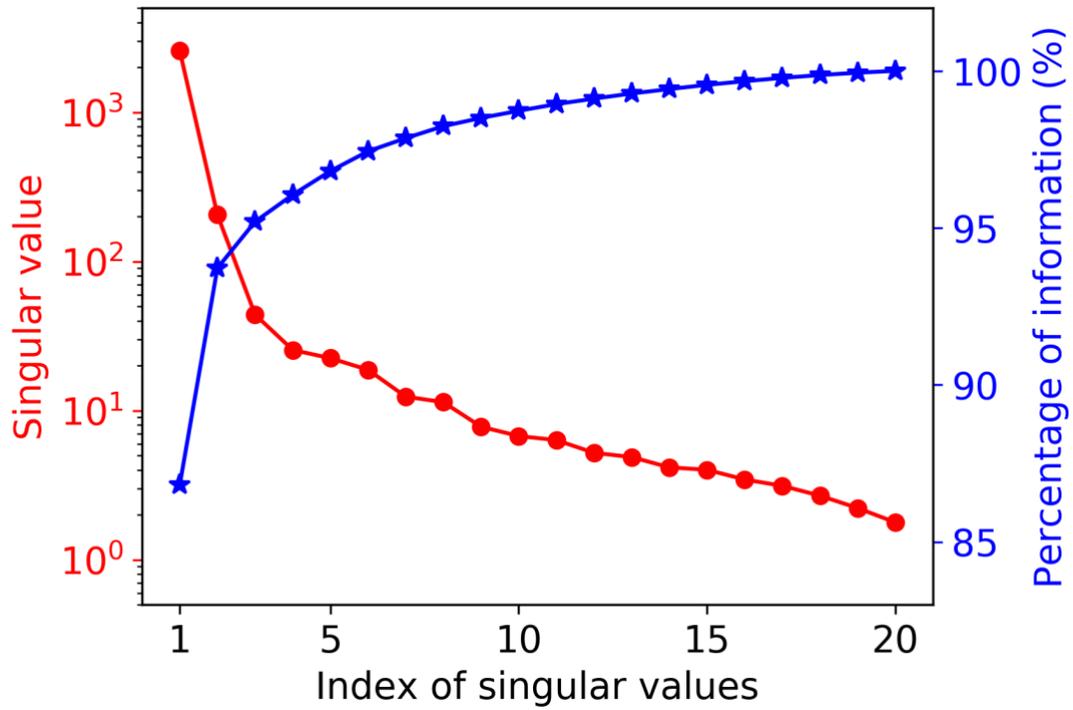

Figure 4. Singular value decay and the information contained in the first largest singular values. The top 5 singular values contain 97% information of training data matrix with 42660 GPP variables and 20 samples.

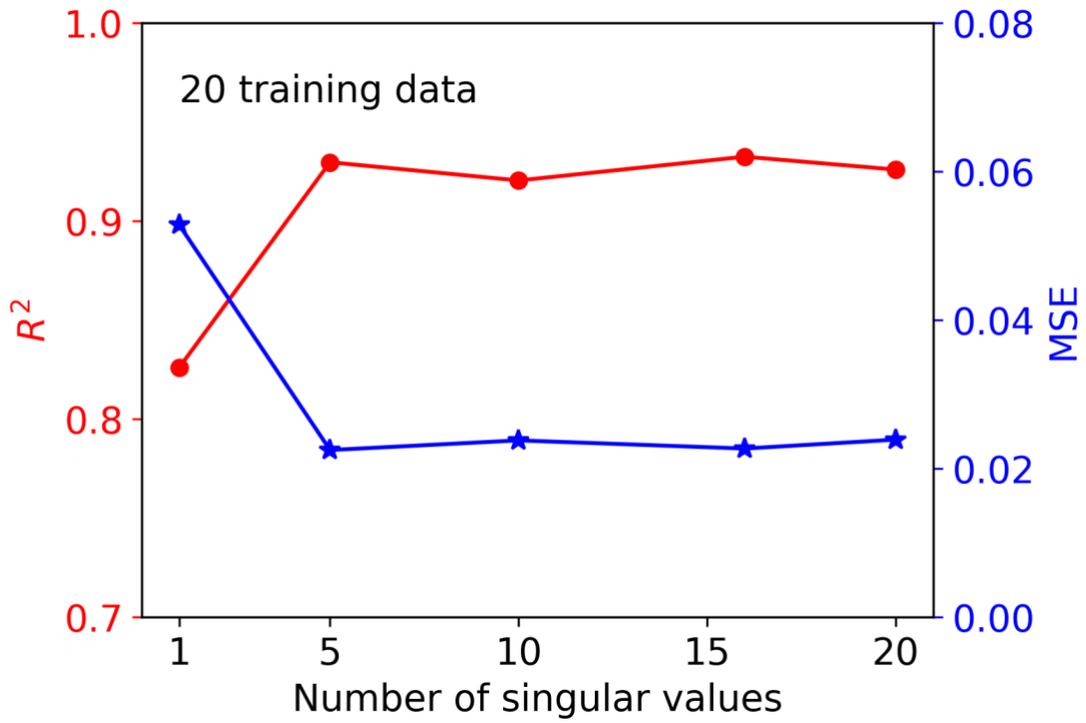

Figure 5. Performance of the NNs trained by 20 data with considering the different number of singular value coefficients after SVD.

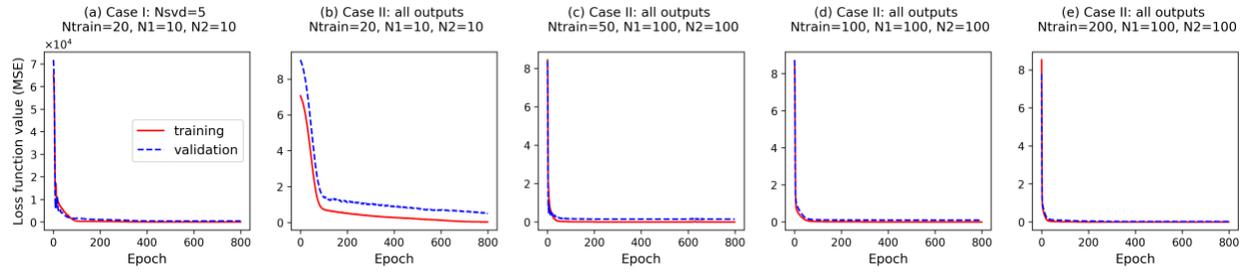

Figure 6. Changes of loss function values along epochs for training and validation data (a) in Case I which builds surrogates of the 5 singular value coefficients with a simple NN (two hidden layers and each layer has 10 nodes, N1=N2=10) based on 20 training data (Ntrain=20), and (b-e) in Case II which builds surrogates of all outputs with different NN architectures and different training data size.

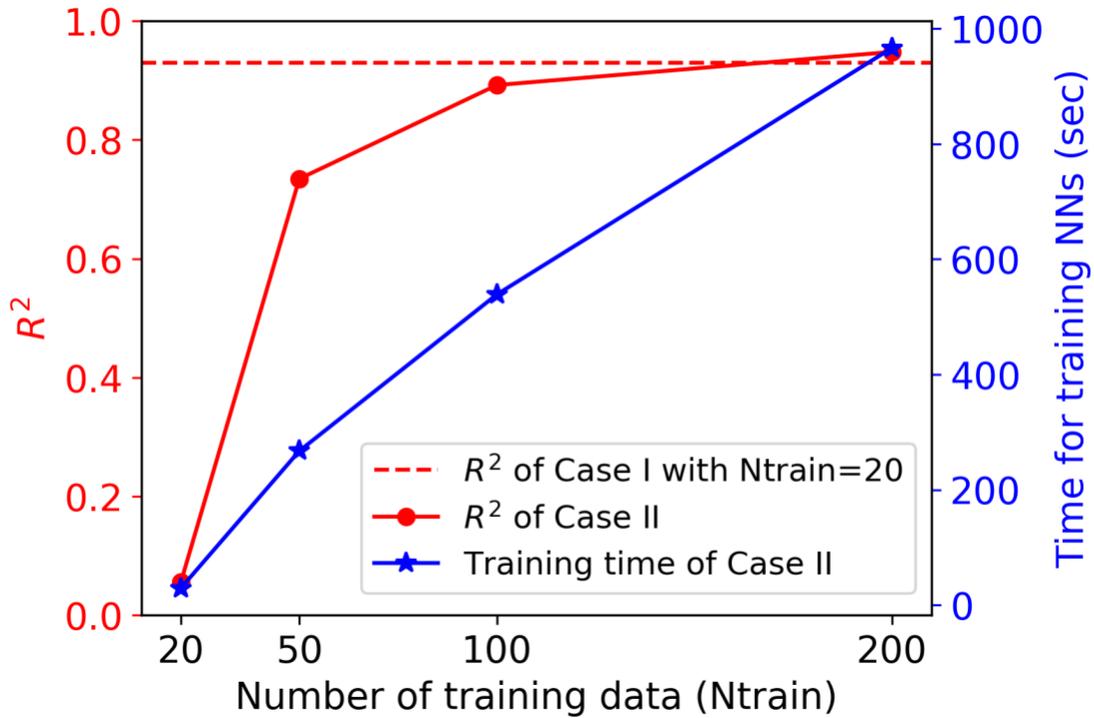

Figure 7. Comparison of NN performance between Case I: building surrogates of 5 singular value coefficients (Nsvd=5) after SVD based on 20 training data (red dashed line) and Case II: building surrogates for all outputs directly with different numbers of training data (red solid line). Each training data represents one sELM simulation. The right y-axis shows the time in training the NN in Case II. The time for training the NN in Case I is 4 seconds.

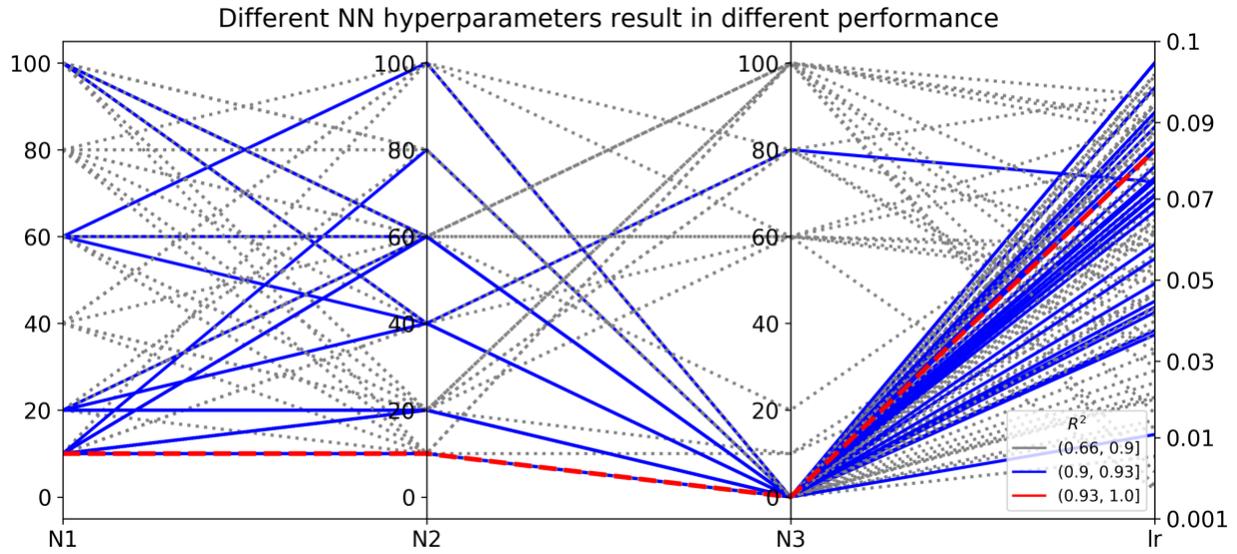

Figure 8. Different sets of NN hyperparameters result in different $R^2$ score in evaluating the 1000 test data. N$l$ is the number of nodes in hidden layer $l$, where $l$=1, 2, and 3. lr is the learning rate of Adam algorithm for NN weights optimization.

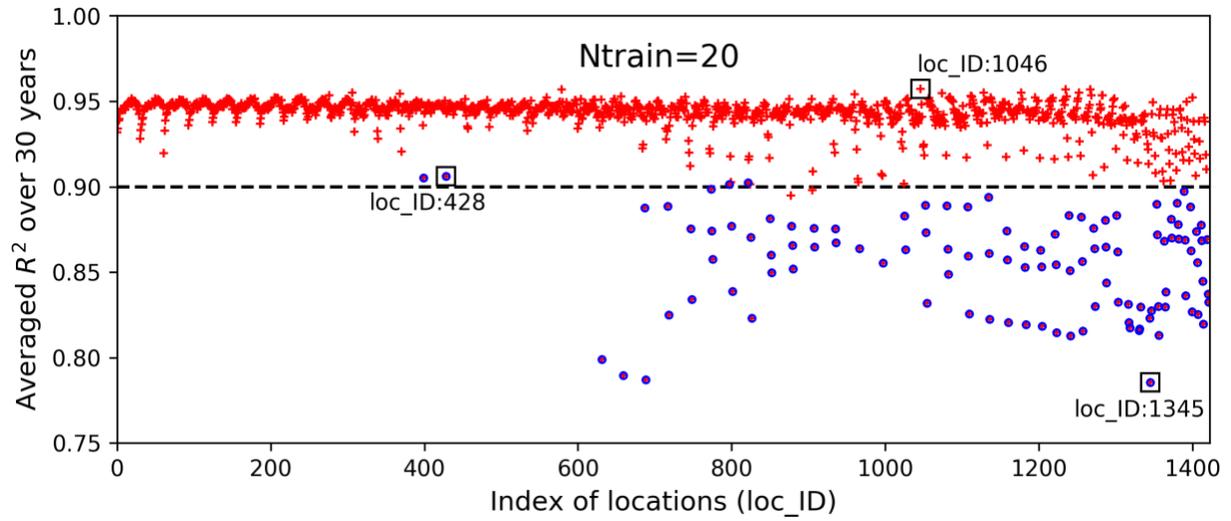

Figure 9. Averaged $R^2$ scores over 30 years at 1422 locations in evaluating the 1000 test data based on 20 training samples, where the blue circles identify the locations having zero GPP simulations.

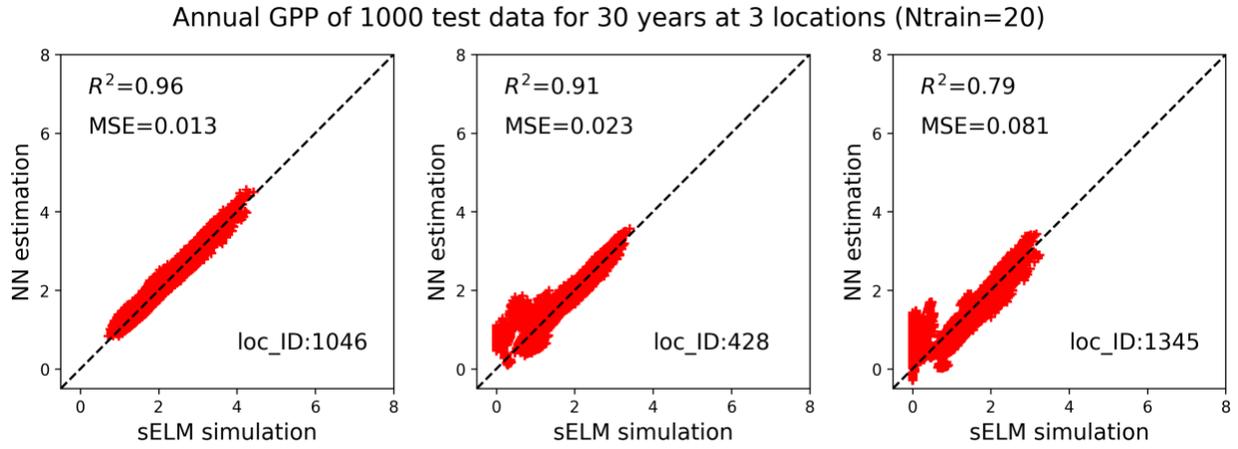

Figure 10. Simulations of annual GPPs (gC/m^2/day) from sELM and NN-based surrogate model in evaluating 1000 test data for 30 years at 3 locations, where the NN is trained by 20 data using our method.

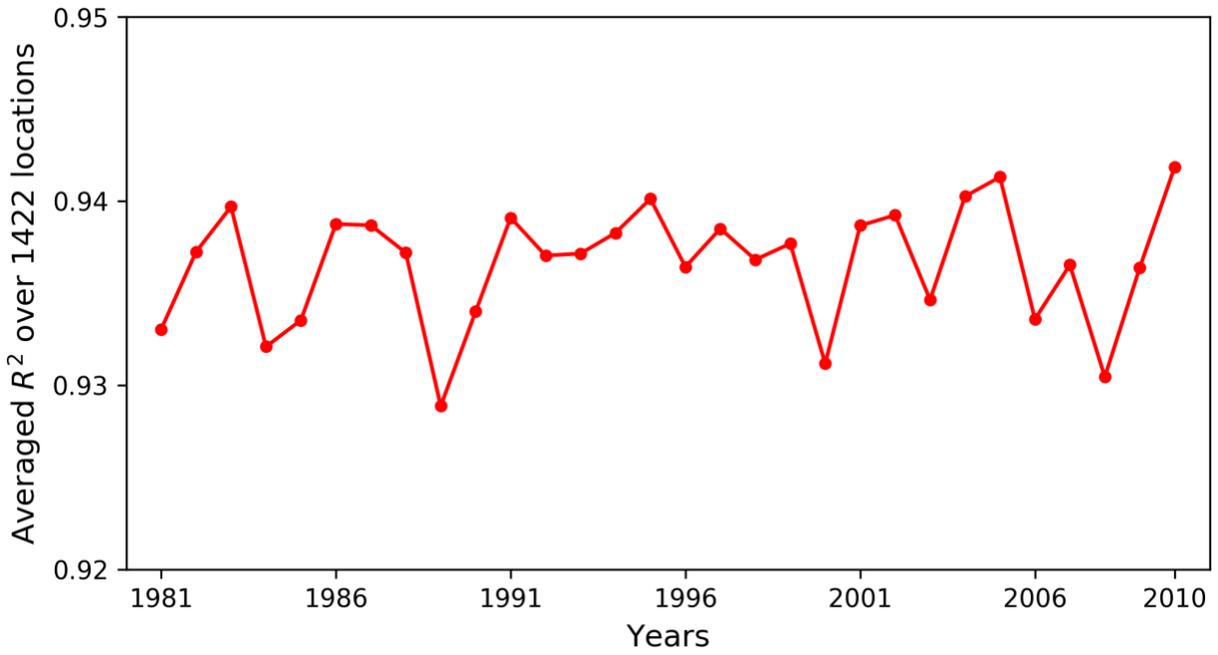

Figure 11. Averaged $R^2$ scores over 1422 locations at 30 years in evaluating the 1000 test data.

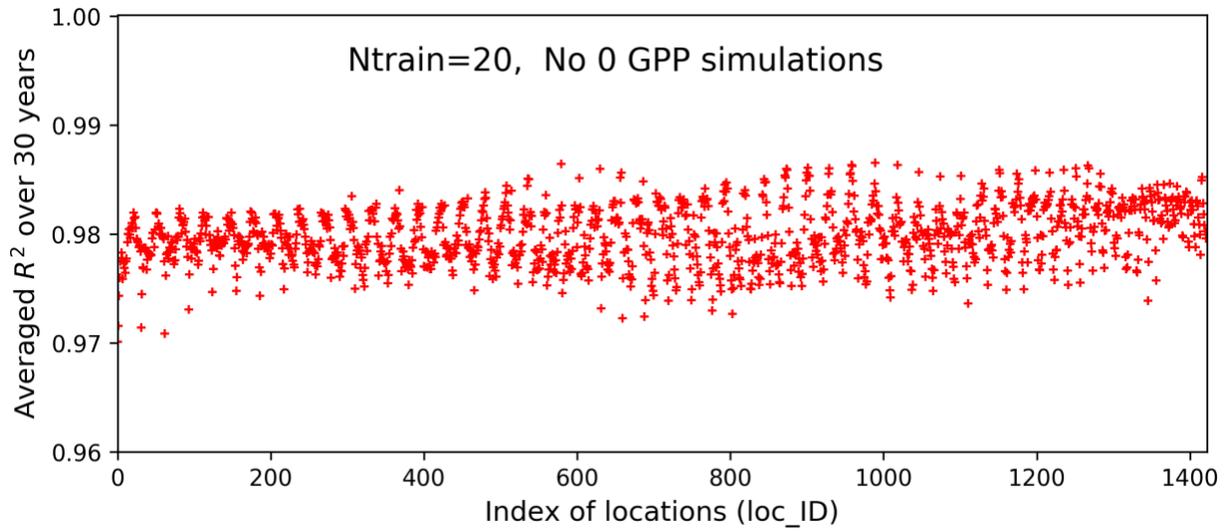

Figure 12. Averaged $R^2$ scores over 30 years at 1422 locations in evaluating the 1000 test data based on 20 training data in experiment I where the samples are generated in a subdomain of the parameter space without zero GPP simulations. The averaged $R^2$ score is 0.98.

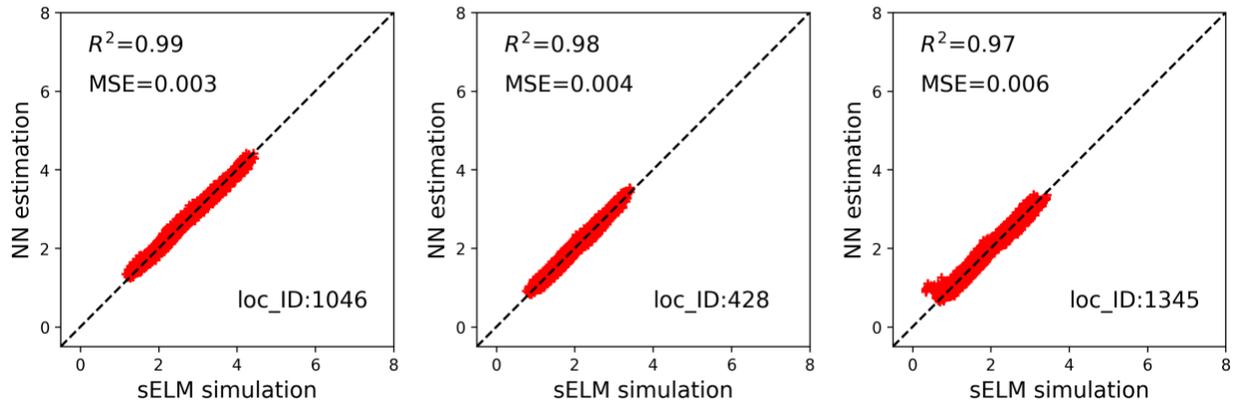

Figure 13. Simulations of annual GPPs (gC/m^2/day) from sELM and NN-based surrogate model in evaluating 1000 test data for 30 years at 3 locations in experiment I where the samples are generated in a subdomain of the parameter space without zero GPP simulations.

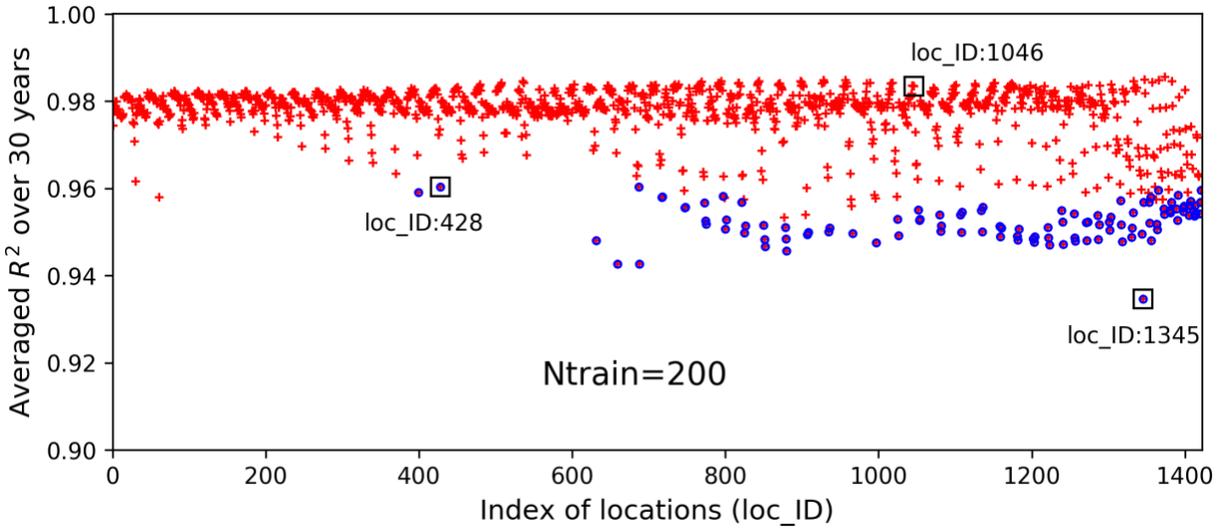

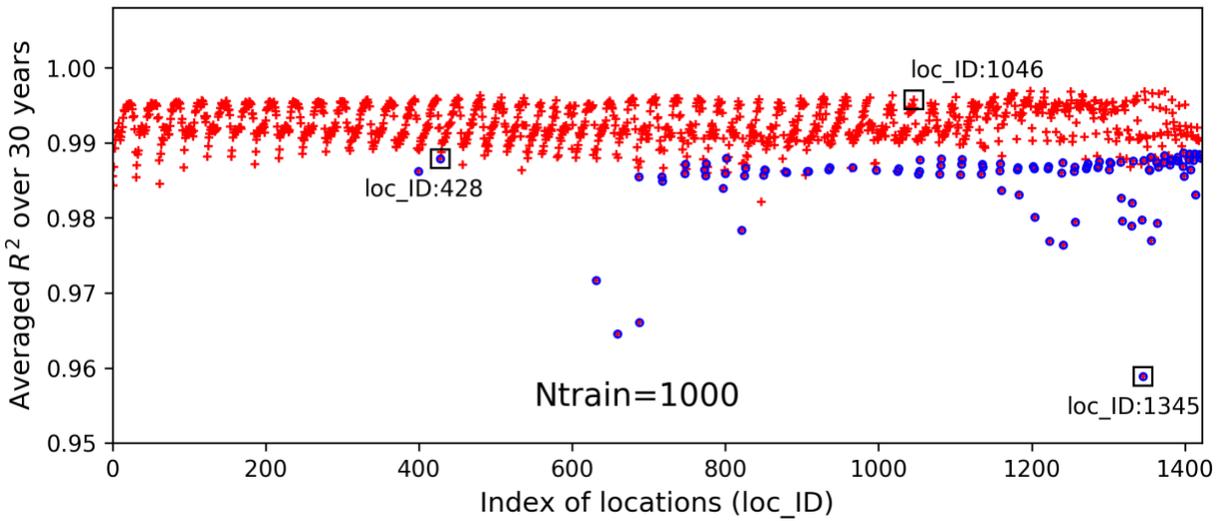

Figure 14. Averaged $R^2$ scores over 30 years at 1422 locations in evaluating the 1000 test data based on 200 and 1000 training samples, where the blue circles identify the locations having zero GPP simulations.

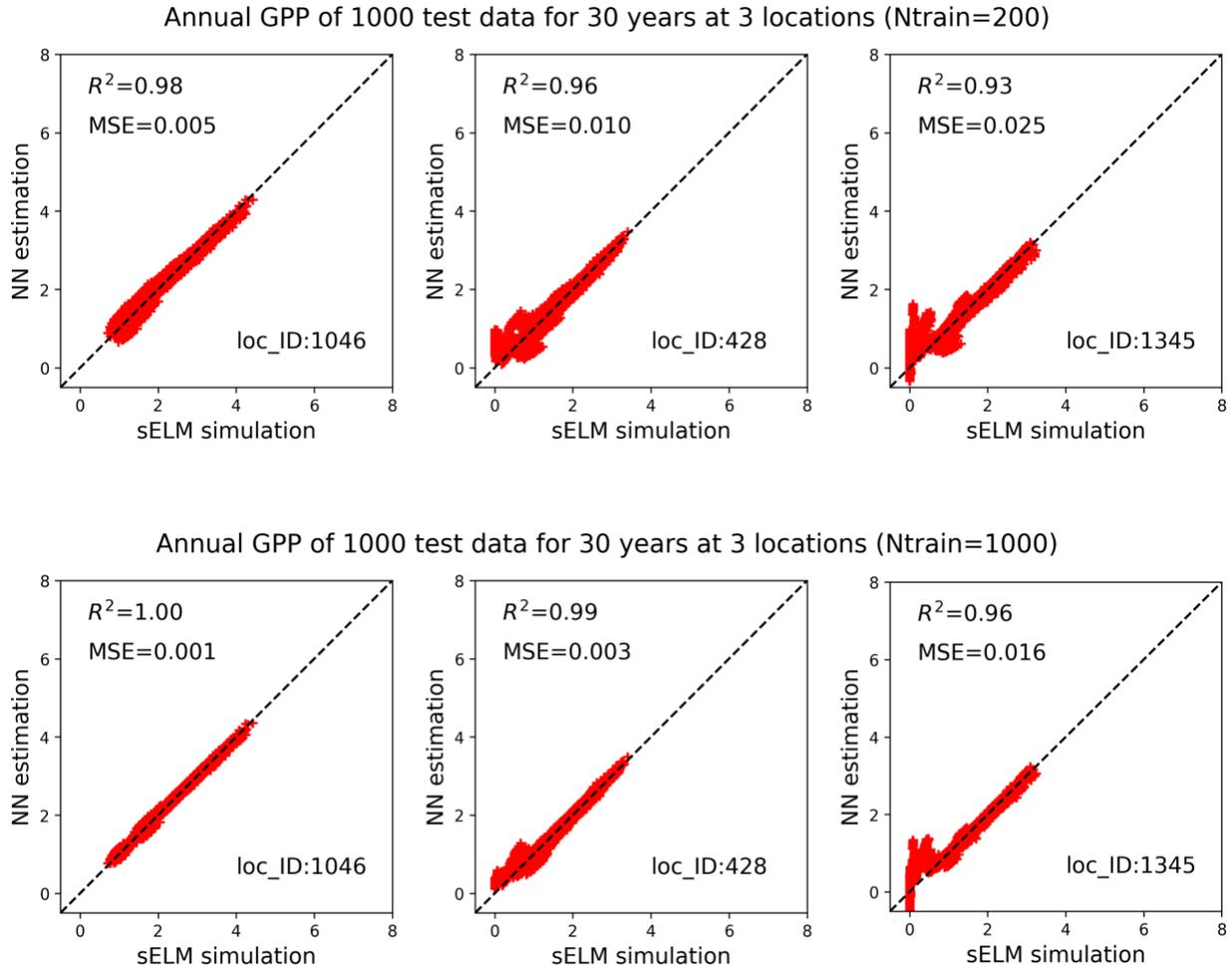

Figure 15. Simulations of annual GPPs (gC/m^2/day) from sELM and NN-baed surrogate model in evaluating 1000 test data for 30 years at 3 locations, where the NN is trained by 200 and 1000 data.